\documentclass{article}

\usepackage{microtype}
\usepackage{graphicx}
\usepackage{subfigure}
\usepackage{booktabs} 
\usepackage{hyperref}

\usepackage[accepted]{icml2018}

\usepackage{amsfonts} 
\usepackage{amsopn}
\usepackage{kotex}
\usepackage[export]{adjustbox}

\DeclareMathOperator*{\argmax}{arg\,max}

\newcommand{\yj}[1]{\textcolor{black}{#1}}
\newcommand{\sm}[1]{\textcolor{black}{#1}}

\newcommand{\nj}[1]{\textcolor{black}{#1}}

\icmltitlerunning{BOOK: Storing Algorithm-Invariant Episodes for Deep Reinforcement Learning}
\begin{document}

\twocolumn[
\icmltitle{BOOK: Storing Algorithm-Invariant Episodes \\ for Deep Reinforcement Learning}
\begin{center}
\icmlauthor{Simyung Chang$^{1,2}$, YoungJoon Yoo$^{3}$, Jaeseok Choi$^{1}$, Nojun Kwak$^{1}$}{}
\\$^{1}$Seoul National University, $^{2}$Samsung Electronics, $^{3}$Clova AI Research, NAVER Corp.
\\\texttt{$^{1}$\{timelighter, jaeseok.choi, nojunk\}@snu.ac.kr, $^{3}$youngjoon.yoo@navercorp.com}
\end{center}



\icmlkeywords{Machine Learning, ICML}

\vskip 0.3in


]




\begin{abstract}
We introduce a novel method to train agents of reinforcement learning (RL) by sharing knowledge in a way similar to the concept of using a book. 
The recorded information in the form of a book is the main means by which humans learn knowledge. 
Nevertheless, the conventional deep RL methods have mainly focused either on \textit{experiential learning} where the agent learns through interactions with the environment from the start or on \textit{imitation learning} that tries to mimic the teacher.
Contrary to these, our proposed \textit{book learning} shares key information among different agents in a book-like manner by delving into the following two characteristic features: 
(1) By defining the linguistic function, input states can be clustered semantically into a relatively small number of core clusters, which are forwarded to other RL agents in a prescribed manner. 
(2) By defining state priorities and the contents for recording, core experiences can be selected and stored in a small container. 
We call this container as `BOOK'. 
Our method learns hundreds to thousand times faster than the conventional methods by learning only a handful of core cluster information, which shows that deep RL agents can effectively learn through the shared knowledge from other agents.
\end{abstract}

\section{Introduction}
\label{sec:introduction}

\begin{figure}[t]
\centering
\subfigure{\includegraphics[trim={0.1cm 0.1cm 0.1cm 0.1cm}, clip, frame, width=0.49\linewidth]{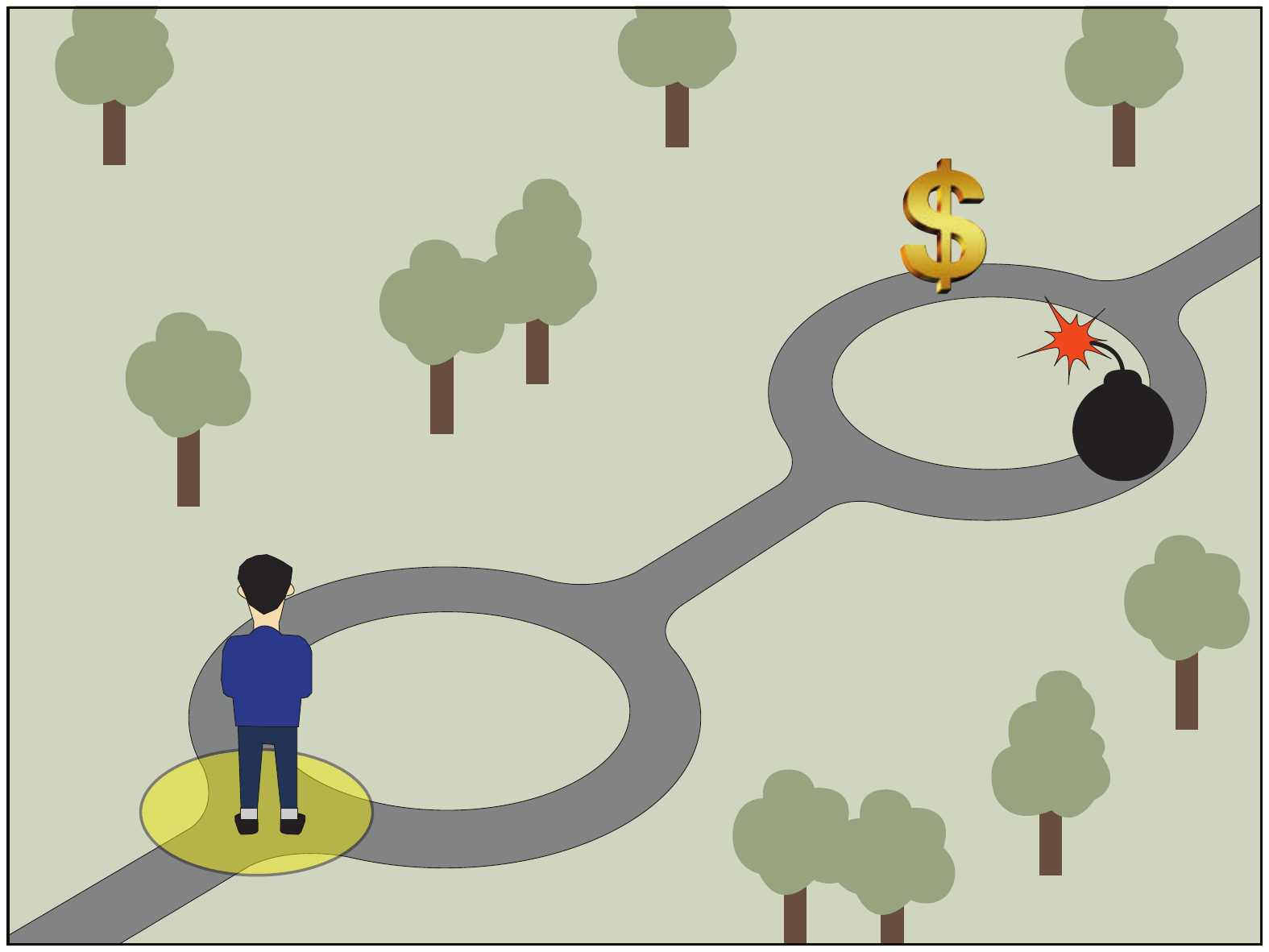}}
\subfigure{\includegraphics[trim={0.1cm 0.1cm 0.1cm 0.1cm}, clip, frame, width=0.49\linewidth]{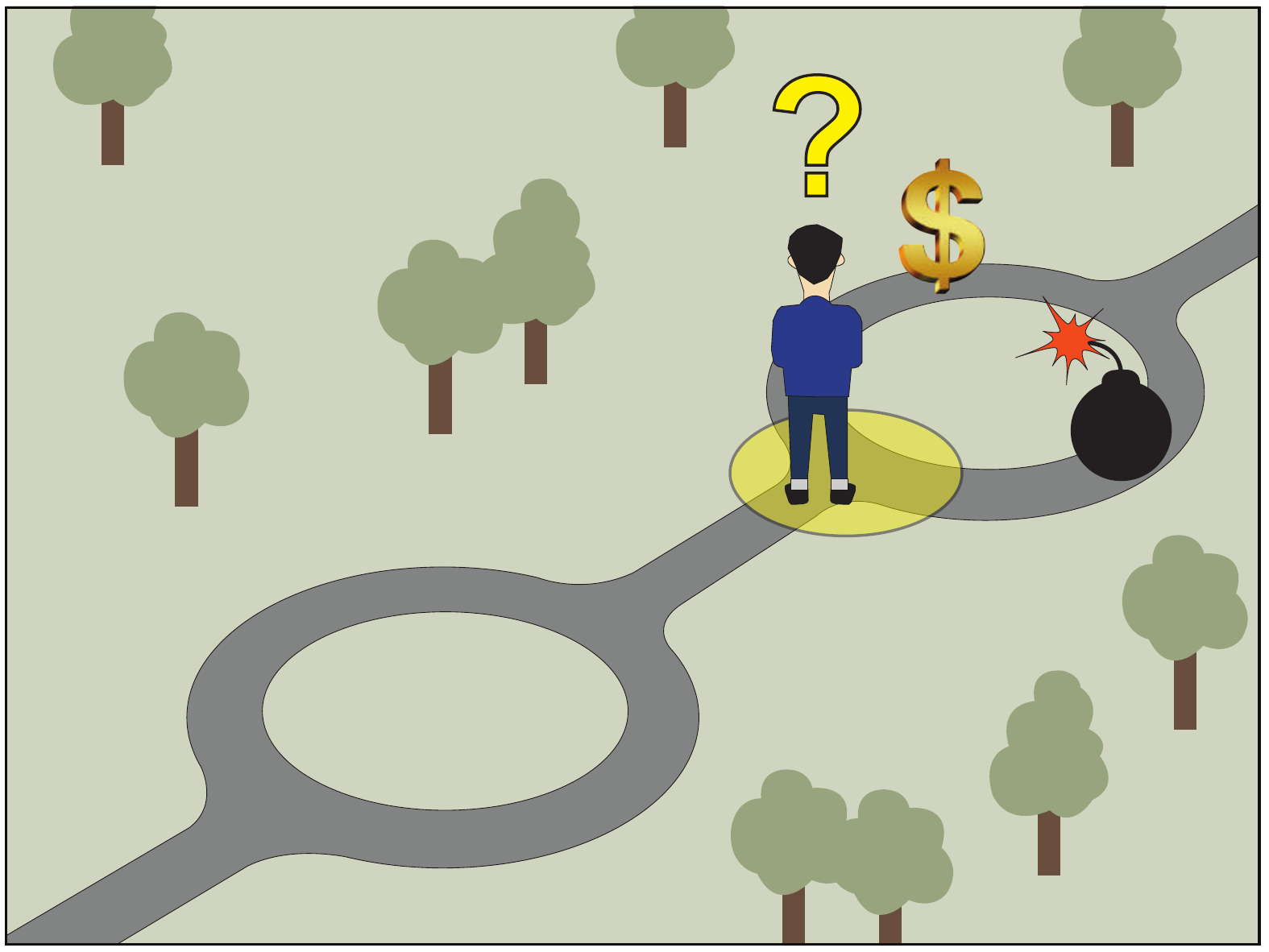}}
\vspace{-2mm}
\caption{
Example illustration of the semantically important state.
In the left image, the person (agent) standing on the yellow circle (state) can choose either ways, and the results for two actions would be the same. 
Conversely, in the right image, the result will be largely different (bomb or money) according to the action (direction) the agent choose on the turning point (yellow).
In our work, the state in the right image is considered to be more important than the state in the left image and this state is stored for further usage in the learning of other agents.
}
\label{fig:Importance}
\end{figure}




Recently, reinforcement learning (RL) using deep neural networks \cite{mnih2013playing,van2016deep,mnih2016asynchronous} has achieved massive success in control systems consisting of complex input states and actions, and applied to various research fields \cite{silver2016mastering,abbeel2007application}.
The RL problem is not easy to directly solve via cost minimization problem because of the constraint that it is difficult to immediately obtain the output according to the input.
Therefore, various methods such as $Q$-learning \cite{bellman1957markovian} and policy gradient \cite{sutton1999policy} have been proposed to solve the RL problems.

The recent neural-network (NN)-based RL methods \cite{mnih2013playing,van2016deep,mnih2016asynchronous} approximate the dynamic-programming-based (DP-based) optimal reinforcement learning ~\cite{jaakkola1994convergence} through the neural network.
However, this process has the problem that the $Q$-values for independent state-action pairs are correlated, which violates the independence assumption. Thus, this process is no longer optimal~\cite{werbos1992approximate}.
This results in differences in performance and convergence time depending on the experiences used to train the network.
Hence, the effective selection of the experiences becomes crucial for successful training of the deep RL framework.

To gather the experiences, most deep-learning-based RL algorithms~\cite{mnih2013playing,van2016deep,mnih2016asynchronous,schulman2015trust,riedmiller2005neural} have utilized experience memory in the learning process~\cite{lin1992self}, which stores batches of state-action pairs (experiences) that frequently appear in the network for repetitive use in the future learning process.
Also,  in \cite {schaul2015prioritized}, a method of prioritized experience memory that finds priorities of each experience is proposed, based on which a batch is created.
Eventually, the key to creating such a memory is to compute the priorities of the credible experiences so that learning can focus on the reliable experiences.
However, in the existing methods, just a few episodes have meaningful information, and the usability of the gathered episodes are highly algorithm-specific.
It can be largely inefficient compared to humans who can select semantically meaningful (credible) events for learning the proper behaviors, regardless of the training method.
Figure \ref{fig:Importance} shows the examples of the case.
In the situation shown in the left image, choosing an action does not bring much difference to the agent. 
However, in the case of the right image, the result (bomb or money) of choosing an action (left or right) can be significantly different for the agent, and it is natural to think that the later case is much important for deciding the movement of the agent.

Inspired by the observation, in this paper, we propose a method of extracting and storing important episodes that 
are invariant to diverse RL algorithms.
First, we propose an importance and a priority measures that can capture the semantically important episodes during entire experiences.
More specifically, in this paper, the importance of a state is measured by the difference of the rewards resultant from different actions and the priority of a state is defined as the product of importance and the frequency of the state in episodes.

Then, we gather experiences during an arbitrary deep RL learning procedure, and store them into dictionary-type memory called `BOOK' (Brief Organization of Obtained Knowledge). The process of generating a BOOK during the learning of a writer agent will be termed as 'Writing the BOOK' in the followings.
The stored episodes are quantized with respect to the state, and the quantized states are used as a key in the book memory.
All the experiences in the BOOK are dynamically updated by upcoming experiences having the same key.
To efficiently manage the episode in the BOOK, some linguistics inspired terms such as linguistic function and state are proposed. 

We have shown that the 'BOOK' memory is particularly effective for two aspects. 
First, we can use the memory as a good initialization data for diverse RL training algorithms, which enables fast convergence. 
Second, we can achieve compatible, and sometimes higher performances by only using the experiences in the memory when training a RL network, compared to the case that entire experiences are used.
The experiences stored in the memory is usually \sm{a few hundred  times} smaller compared to the experiences required in usual random-batch-based RL training~\cite{mnih2013playing,van2016deep}, and hence give us much effectiveness in time and memory space required for the training.

The contributions of the proposed method are as follows: 

(1) The dictionary termed as BOOK that stores the credible experience, which is useful for diverse RL network training algorithms, expressed by the tuple (cluster of states, action, and the corresponding $Q$-value) is proposed.

(2) The method for measuring the credibility: importance and priority terms of each experience valid for arbitrary RL training algorithms, is proposed.

(3) The training method for RL that utilizes the BOOK is proposed, which is inspired by DP and is applicable to diverse RL algorithms.

To show the efficiency of the proposed method, it is applied to the major deep RL methods such  DQN~\cite{mnih2013playing} and A3C~\cite{mnih2016asynchronous}. 
The qualitative as well as the quantitative performances of the proposed method are validated through the experiments on public environments published by OpenAI~\cite{brockman2016openai}.


\section{Background}
\label{sec:background}

\begin{figure*}[t]
    \centering
    \includegraphics[width=0.99\linewidth]{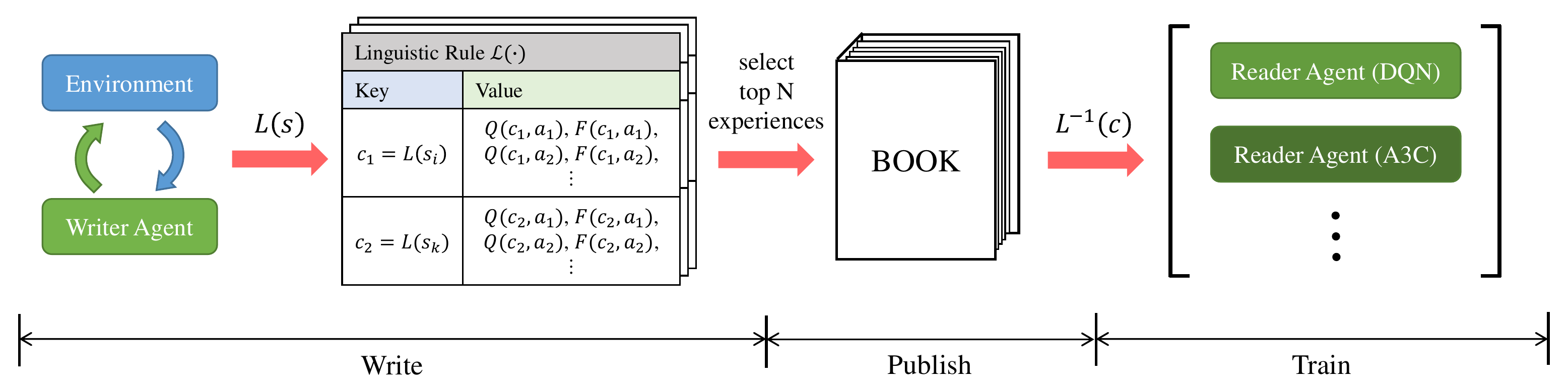}
    \vspace{-4mm}
    \caption{
    Overall framework of the proposed model. Similar experiences (state-action pairs) from multiple episodes are grouped into a cluster and the credible experiences corresponding to large clusters are written in a BOOK with their $Q$-values and frequencies $F$s. The BOOK is published with Top N experiences after learning. Then, reader agents use this information in training. 
}
    \label{fig:framework}
    \vspace{-2mm}
\end{figure*}

The goal of RL is to estimate the sequential actions of an agent that maximize cumulative rewards given a particular environment.
In RL, Markov decision process (MDP) is used to model the motion of an agent in  the environment.
It is defined by the \textit{state} $s_t\in\mathbb{R}^S$, \textit{action} $a_t \in\{a_1,\ldots,a_A\}$ which occurs in the state $s_t$, and the corresponding \textit{reward} $r_t \in \mathbb{R}$, at a time step $t \in \mathbb{Z}^+$. \footnote{$\mathbb{R}$ and $\mathbb{Z}^+$ denote the real and natural numbers respectively.} 
We term the function that maps the action $a_t$ for a given $s_t$ as the \textit{policy}, and the future state $s_{t+1}$ is defined by the pair of the current state and the action, $(s_t, a_t)$.
Then, the overall cost for the entire sequence from the MDP is defined as
the accumulated discounted reward, $R_t = \sum_{k=0}^\infty \gamma^k r_{t+k}$, with a discount factor $\gamma \le 1$.
 

Therefore, we can solve the RL problem by finding the optimal policy that maximizes the cost $R_t$.
However, it is difficult to apply the conventional optimization methods in finding the optimal policy.
It is because we should wait until the agent reaches the terminal state to see the cost $R_t$ resulting from the action of the agent at time $t$.
To solve the problem in a recursive manner, we define the function $Q(s_t,a_t)=\mathbb{E}[R_t|s=s_t,a=a_t,\pi]$ denoting the expected accumulated reward for $(s_t,a_t)$ with a policy $\pi$.
Then, 
we can induce the recurrent Bellman  equation~\cite{bellman1957markovian}:
\begin{equation}
Q(s_t,a_t) = r_{t}+\gamma\max_{a_{t+1}}Q(s_{t+1},a_{t+1}).
\label{eq:q_eqn}
\end{equation}
It is proven that the $Q$-value, $Q(s_t,a_t)$, for all time step $t$ satisfying (\ref{eq:q_eqn}) can be calculated by applying dynamic programming (DP), and the resultant $Q$-values are optimal~\cite{jaakkola1994convergence}.
However, it is practically impossible to apply the DP method when the number of state is large, or the state is continuous.
Recently, the methods such as Deep $Q$-learning (DQN), Double Deep-$Q$-learning (DDQN) solve the RL problem with complex state $s_t$ by using approximate DP 
that trains $Q$-network.
The $Q$-network is designed so that it calculates the $Q$-value for each action when a state is given.
Then, the $Q$-network is trained by the temporal difference (TD) \cite{watkins1992q} method which reduces the gap between $Q$-values acquired from the $Q$-network and those from (\ref{eq:q_eqn}).

\section{Related Work}
\label{sec:related_work}

Recently, deep learning methods~\cite{mnih2013playing, hasselt2010double, van2016deep, wang2015dueling, mnih2016asynchronous, schaul2015prioritized,salimans2017evolution} have improved performance by incorporating neural networks to the classical RL methods such as Q-learning~\cite{watkins1992q}, SARSA~\cite{rummery1994line}, evolution learning~\cite{salimans2017evolution}, and policy searching methods~\cite{williams1987class, peters2003reinforcement} which use TD~\cite{sutton1988learning}.

\citet{mnih2013playing}, \citet{hasselt2010double} and \citet{van2016deep} replaced the value function of Q-learning with a neural network by using a TD method. \citet{wang2015dueling} proposed an algorithm that shows faster convergence than the method based on Q-learning by applying dueling network method~\cite{harmon1995advantage}. Furthermore, \citet{mnih2016asynchronous} applied the asynchronous method to Q-learning, SARSA, and Advantage Actor-Critic models.


The convergence and performance of deep-learning-based methods are greatly affected by input data which are used to train an approximated solution~\cite{bertsekas1995neuro} of  classical RL methods. \citet{mnih2013playing} and \citet{van2016deep} solved the problem by saving experience as batch in the form of \textit{experience replay memory}~\cite{lin1993reinforcement}. In addition, Prioritized Experience Replay~\cite{schaul2015prioritized} achieved higher performance by applying replay memory to recent Q-learning based algorithms by calculating priority based on the importance of experience.
\yj{\citet{pritzel2017neural} proposed a Neural episodic control (NEC) to apply tabular based Q-learning method for training the Q-network by first, semantically clustering the states and then, updates the value entities of the clusters.} 

Also, imitation learning~\cite{ross2014reinforcement, krishnamurthy2015learning, chang2015learning} which solves problems through 
expert's experience is one of the main research flows. This method trains a new agent in a supervised manner using state-action pairs obtained from the expert agent and shows faster convergence speed and better performance using experiences of the expert. However, it is costly to gather experiences from experts.



\yj{The goal of our work is different to the mentioned approaches} as follows.
(1) compared to imitation learning, the proposed method differs in the aspect that credible data are extracted from the past data in an unsupervised manner, and more importantly, (2) compared to the prioritized experience replay~\cite{schaul2015prioritized},
\yj{our work proposes a method to generate a memory that stores \sm{core} experiences useful for training diverse RL algorithms.
(3) Also, compared to the NEC~\cite{pritzel2017neural}, our work aims to use the BOOK memory for good initialization and fast convergence, when training the RL network \nj{regardless of the algorithm used.}} \sm{but, the dictionary of NEC can not provide all \nj{the} information necessary for learning, such as states,  so it is difficult to use it to train other RL networks.} 


\section{Proposed Method}
\label{sec:proposed_method}

In this paper, our algorithm aims to find the core experience through many experiences and write it into a BOOK, 
which can be used to share knowledge with other agents 
that possibly use different RL algorithms.
%
Figure \ref{fig:framework} describes the main flow of the proposed algorithm. 
First, from the RL network, the terminated episodes of a writer agent are extracted. 
%
Then, among experiences from the episodes, the core and credible experiences are gathered and stored into the BOOK memory.
%
In this process, using the semantic cluster of states as a key, the BOOK stores the value information of the experiences related to the semantic cluster. 
%
This `writing' process is iterated until the end of training.
Then, the final BOOK is `published' with the top $N$ core experiences of this memory, that can be directly exploited in the `training' of other reader RL agents.

In the following subsections, how to design the BOOK and how to use BOOK in the training of RL algorithms are described in more detail.

\subsection{Desigining the BOOK Structrue}

Given a state $s\in \mathbb{R}^S$ and action $a\in\{a_1,...,a_A\}$, we define the memory $\mathcal{B}$ termed as `BOOK' which stores the credible experience in the form appropriate for lookup-table inspired RL.
Assuming there exists semantic correlation among states, the input state $s_i, i=1\ldots N_s$ can be clustered into the core $K$ clusters $\mathcal{C}_k\in\mathcal{C},k=1,\ldots,K$.
To reduce the semantic redundancy, 
the BOOK stores the information related to the cluster $\mathcal{C}_k$, and the corresponding information is updated by the information of the states $s_i$ included in the cluster.
It means that the memory space of the BOOK in the `writing' process is $\mathcal{O}(AK)$.
To map the state $s_i$ to the cluster $\mathcal{C}_k$, we define the mapping function $L:s \rightarrow c_k$, where $c_k\in\mathbb{R}^S$ denotes the representative value of the cluster $\mathcal{C}_k$.
We term the mapping function $L(\cdot)$, the representative $c_k$, and the reward of $c_k$ as \textit{linguistic function}, \textit{linguistic state}, and \textit{linguistic reward}, respectively \footnote{The term `linguistic' is used to represent both characteristics of `abstraction' and `shared rule'.}.
To cluster the states and define the \textit{linguistic function}, arbitrary clustering methods or quantization can be applied. For simplicity, we adopt the quantization in this paper.

Consequently, the element of a BOOK $b_{k,j}\in \mathcal{B}$ is defined as $b_{k,j} \in \{c_k, Q(c_k,a_j), F(c_k, a_j)\}$, where
$Q(c_k,a_j)$ and $F(c_k, a_j)$ denote the $Q$-value of $(c_k,a_j)$ and the hit frequency of the $b_{k,j}$.  
Then, the information regarding the input state $s_i$ is stored into $b_k = [b_{k,1},\ldots, b_{k,A}]$, where $c_k = L(s_i)$.
The $Q$-value $Q(c_k,a_j)$ is iteratively updated by $Q^{t}(s_t = s_i,a_t = a_j)$ which denotes the $Q$-value from the credible experience $\{s_t,a_t,r_{t},s_{t+1}\}$.





\label{subsec:design}
\subsection{Iterative Update of the BOOK using Credible Experiences}

\begin{algorithm}[t]
\begin{algorithmic}
\STATE Define linguistic function $L$ for states and reward and initialize it.
\STATE Initialize BOOK $\mathcal{B}$ with capacity $K$.
\FOR{episode $=1,\ldots, M$}
	\STATE Initialize Episode memory $\mathcal{E}$
    \STATE Get initial state s
    \FOR {$t=t_{start}, \ldots, t_{terminal}$}
    	\STATE Take action $a_t$ with policy $\pi$
		\STATE Receive new state $s_{t+1}$ and reward $r_t$
		\STATE Store transition $\left(s_t,a_t,r_t,s_{t+1}\right)$ in $\mathcal{E}$
		\STATE Perform general reinforcement algorithm 
	\ENDFOR
	\FOR {$t=t_{terminal}, \ldots, t_{start}$}
    	\STATE Take transition $\left(s_t,a_t,r_t,s_{t+1}\right)$ from $\mathcal{E}$
    	\STATE $c_k = L(s_t)$
        \STATE Update $Q(c_k,a_t)$, $F(c_k,a_t)$ to $\mathcal{B}$ with equation (\ref{eq:update_q}), (\ref{eq:def_q}), (\ref{eq:beta}), (\ref{eq:freqency})
    \ENDFOR
\IF {$episode \% T_{decayPeriod}==0$}
\STATE Decay $F(c_k,a_j)$ in $\mathcal{B}$ for all $k \in \{1, \ldots, K \}$ and $j \in \{1,\ldots, A\}$ 
\ENDIF
\ENDFOR
\end{algorithmic}
\caption{Writing a BOOK}
\label{alg:book}
\end{algorithm}

To fill the BOOK memory by credible experiences, we first extract the credible experiences from the entire possible experiences.
We extract the credible experiences based on the observation that the terminated episode\footnote{An episode denotes a sequence of state-action-reward until termination.} holds valid information to judge whether an agent's action was good or bad. 
At least in the terminal state, we can evaluate whether the state-action pair performed good or bad 
by just observing the result of the final action; for example, success or failure.
Once we get credible experience from terminal sequences, then we can get the related credible experiences using the upcoming equation (\ref{eq:update_q}).
More specifically, the BOOK is updated using the experience $E_t$ from the terminated episode $\mathcal{E} = \{E_1,...,E_T\}$ in backward order, i.e., from $E_T$ to $E_1$, where $E_t = \{s_t,a_t,r_{t},s_{t+1}\}$. 
%
%
Consider that for an experience $E_t$ at time $t$, the current state, current action, and the future state are $s_t = s_i, a_t = a_j$ and $s_{t+1} = s_{i'}$, respectively. Also, assume that $s_i \in \mathcal{C}_k$. 
Then, the $Q$-value $Q(c_k, a_j)$ stored in the content $b_{k,j}$ is updated by 
\begin{equation}
Q(c_k,a_j) = \beta Q(c_k,a_j) + (1-\beta)Q^t(s_i,a_j),
\label{eq:update_q}
\end{equation}
where
\begin{equation}
    Q^{t}(s_i, a_j) = r_{t} + \gamma \max_{a'}Q(s_{i'}, a').
\label{eq:def_q}
\end{equation}
\begin{equation}
\beta = F(c_k,a_j)/\{F(c_k,a_j)+F(L(s_{i'}),\argmax_{a'}Q(s_{i'}, a'))\}.
\label{eq:beta}
\end{equation}
Here, $F(c_k,a_j)$ refers to the hit frequency of the content $b_{k,j}$. 
The term $Q^t(s_i,a_j)$ denotes the estimated $Q$-value of $(s_i,a_j)$ acquired from the RL network.
In (\ref{eq:beta}),  $F(L(s_{i'}),\argmax_{a'}Q(s_{i'}, a'))$ is initialized to $1$ when the term regarding $L(s_{i'})$ is not yet stored in the BOOK.
We note that we calculate $Q^{t}(s_t, a_t)$ from $Q^{t}(s_{t+1}, a_{t+1})$ in backward manner, because only the terminal experience $E_T$ is fully credible among the episode $\mathcal{E}$ acquired from the RL network.
The update rule for the frequency term $F(c_k,a_j)$ in the content $b_{k,j}$ is defined as 
\begin{small}
\begin{equation}
F(c_k,a_j) = \min(F(c_k,a_j)+F(L(s_{i'}),\argmax_{a'}Q(s_{i'}, a'),F_{l}),
\label{eq:freqency}
\end{equation}
\end{small}
where $F_{l}$ is the predefined limit of the frequency $F(\cdot,\cdot)$. 
The frequency $F(\cdot,\cdot)$ is reduced by 1 for every predefined number of episodes to avoid $F(\cdot,\cdot)$ from being continually increasing.
To extract the episode $\mathcal{E}$, we can use arbitrary deep RL algorithm based on $Q$-network. Algorithm \ref{alg:book} summarizes the procedure of writing a BOOK.


\label{subsec:update}

\subsection{Priority Based Contents Recoding}

In many cases, the number of clusters becomes large, and it is clearly inefficient to store all the contents without considering the priority of a cluster.
Hence, we maintain the efficiency of BOOK by continuously removing contents with lower priority from the BOOK.
In our method, the priority $p_{k,j}$ is defined by the product of the frequency term $F(c_k,a_j)$ and the importance term $I(c_k)$,
\begin{equation}
p_{k,j} = I(c_k) F(c_k,a_j).
\label{eq:priority}
\end{equation}
The importance term $I(c_k)$ reflects the maximum gap of reward for choosing an action for a given linguistic state $c_k$, as the following:
\begin{equation}
I(c_k) = \max_{a}Q(c_k,a) - \min_{a}Q(c_k,a).
\label{eq:importance}
\end{equation}
In Fig.~\ref{fig:Importance}, we can see the concept of the importance term.
At the first crossroad (state) in the left, the penalty of choosing different branches (actions) is not severe. However, at the second crossroad, it is very important to choose a proper action given the state.
Obviously, the situation in the right image is much crucial, and the RL should train the situation more carefully. Now, we can keep the size of the BOOK as we want by eliminating the contents with lower priority $p_{k,j}$ (left image in the figure).
\label{subsec:priority}

\subsection{Publishing a BOOK}
We have seen how to write a BOOK in the previous subsections. In the `writing' stage in Fig. \ref{fig:framework}, it limits the contents to be kept according to priority, but maintains a considerable capacity $K$ to compare information of various states. 
However, our method finally publish the BOOK with only the top $N (< K)$ priority states with the same rule as the subsection \ref{subsec:priority} after learning of the writer agent.
We have shown through experiments that we can obtain good performance even if a relatively small-sized BOOK is used for training. See section~\ref{sec:Experiment} for more detailed analysis.

\label{subsec:publish}

\subsection{Training Reader Network using the BOOK}






As shown in Figure~\ref{fig:framework}, we train the RL network using the BOOK structure that stores the experience from the episode. 
The BOOK records the information of the representative states that is useful for RL training. 
The information required to learn the general reinforcement learning algorithm can be obtained in the form of $(s, a, Q(s,a))$ or $(s, a, V(s), A(s,a))$ through our recorded data. Here, $V(s)$ and $A(s,a)$ are the value of the state $s$ and the advantage of the state-action pair $(s,a)$.

To utilize the BOOK in the learning of the environment, the linguistic state $c_k$ has to be converted to the real state $s$. The state $s$ can be decoded by implementing the inverse function $s = L^{-1}(c_k)$, or one of the state $s \in \mathcal{C}_k$ can be stored in the BOOK as a sample when the BOOK is made.


In the first case of using $Q$-value $Q(s,a)$ in the training, the recorded information can be used as it is. 
In the second case, $V(s)$ is calculated as the weighted sum of the $Q(s,a)$ and the difference between the $Q$-value and the state value $V$ is used as the advantage $A(s,a)$ as follows: 
\begin{equation}
V(s) \approx \frac{\sum\nolimits_{a_i}F(c_k,a_i)Q(c_k,a_i)}{\sum\nolimits_{a_i}F(c_k,a_i)},
\label{eq:valuefrombook}
\end{equation}
\begin{equation}
\qquad
A(s,a) \approx Q(c_k,a) - V(s).
\label{eq:advantagefrombook}
\end{equation}
%
A BOOK stores only the measured (experienced) data regardless of the RL model without bootstrapping. 
The learning method of each model is used as it is, in the training using the BOOK.
Since DQN~\cite{mnih2013playing} requires state, action and Q-value in learning, it learns by decoding this information in the BOOK. On the other hand, A3C~\cite{mnih2016asynchronous} and Dueling DQN~\cite{wang2015dueling} require state, action, state-value $V$ and advantage $A$, so these decode the corresponding information in the BOOK as shown in equations (\ref{eq:valuefrombook}) and (\ref{eq:advantagefrombook}). \sm{\nj{Because} a BOOK has all the information needed to train \nj{an} RL agent, the agent is not required to interact with the environment while learning the BOOK.}

We note that our learning process shares the essential philosophy with the classical DP in that the learning process explores the state-action space based on credible $Q(s,a)$ stored in the BOOK without bootstrapping and dynamically updates the values in the solution space using the stored information.
As verified by the experiments, we confirmed that our methods achieved better performance with much smaller iteration compared to the existing approximated 
DP-based RL algorithms~\cite{mnih2013playing,mnih2016asynchronous}.

\label{subsec:train}

\section{Experiments}
\label{sec:Experiment}
\begin{table*}[tb]
\caption{Performance of BOOK based learning. \textbf{Score}: An average score that can be obtained by learning the BOOK of size 1,000.
\textbf{Transition}: the number of timesteps that is needed for each reader model to get the same ‘Score’ without learning a BOOK.
\textbf{Ratio}: a ratio of the size of a BOOK over Transition,  Ratio = the size of BOOK / Transition.
}
\label{table:book_exp}
\vskip 0.05in
\begin{center}
\begin{small}
\begin{sc}
\resizebox{1.0\linewidth}{!}{
\begin{tabular}{c|c|c@{\hskip4pt}c@{\hskip4pt}c|c@{\hskip4pt}c@{\hskip4pt}c|c@{\hskip4pt}c@{\hskip4pt}c|c@{\hskip4pt}c@{\hskip4pt}c}
\toprule
\multicolumn{2}{c|}{Model} & \multicolumn{3}{c|}{Cartpole} & \multicolumn{3}{c|}{Acrobot} & \multicolumn{3}{c|}{Lunar Lander} & \multicolumn{3}{c}{Q*bert} \\
\midrule
Writer & Reader & Score  & Transition  & Ratio  & Score  & Transition  & Ratio & Score  & Transition& Ratio  & Score & Transition & Ratio \\
\midrule
  & DQN  & 428.3  & 352K & 0.28\% & \textbf{-280.7} & 13K & 7.7\% & -251.7 &  9.7K & 10.3\% & 196.9 & 566K & 0.18\% \\
DQN & A3C & 500.0 & 324K & 0.31\% & -281.6 & 162K & 0.62\% & -178.2 & 337K & 0.30\% & 302.5 & 182K & 0.55\% \\
  & Dueling & 500.0 & 624K & 0.16\% & -370.2 & 10K & 10.0\% &\textbf{-127.4}& 12K & 8.3\%  & \textbf{324.6} & 931K & 0.11\% \\
\midrule      
& DQN   & 500.0  & 792K & 0.13\% & -172.1 & 49K & 2.0\% & -241.4 & 9.9K & 10.1\% & 290.0  & 880K & 0.11\% \\
A3C & A3C  & 500.0 & 324K & 0.31\% & \textbf{-91.8} & 372K & 0.27\% & \textbf{-144.9} & 520K & 0.19\% & \textbf{436.0} & 383K & 0.26\% \\
  & Dueling & 500.0 & 624K&0.16\% & -177.9  & 32K & 3.1\% &-160.5& 10K & 10.0\%  & 388.1 & 1,080K & 0.09\% \\ 
\bottomrule  
\end{tabular}
}
\end{sc}
\end{small}
\end{center}
\vskip -0.1in
\end{table*}
To show the effectiveness of the proposed concept of BOOK, we tested our algorithm on 4 problems from 3 domains. These are carpole \cite{barto1983neuronlike}, acrobot~\cite{geramifard2015rlpy}, Box2D \cite{catto2011box2d} lunar lander, and Q*bert from Atari 2600 games.
All the experiments were performed using OpenAI gym~\cite{brockman2016openai}. 

The purpose of the experiments is to answer the following questions:
(1) Can we effectively represent valuable information for RL among the entire state-action space and find important states? If so, can this information be effectively transfered to train other RL agent? 
(2) Can the information generated in this way be utilized to train the network in different architecture? For example, can a BOOK generated by DQN be effectively used to train A3C network?

\subsection{Performance Analysis}
In these experiments, we first trained the conventional network of A3C or DQN. During the training of the conventional writer network, a BOOK is written. Then, we tested the effectiveness of this BOOK with two different scenarios. First, we trained the RL networks using only the contents of the BOOK as described in Section \ref{subsec:train}. 
For the second scenario, we conducted additional training for the RL networks that are already trained using the BOOK at first scenario.

\textbf{Performance of BOOK based learning: }
Table \ref{table:book_exp} shows the performance when training the conventional RL algorithm with only the contents of the BOOK. The BOOK is written in the training of writer network with DQN and A3C and published in size of 1,000. Then, reader networks were trained with this BOOK using several different algorithms such as DQN, A3C, and Dueling DQN. This normally took much less time (less than 1 minute in all experiments) than the training of the conventional network from the start without utilizing BOOK. Then, we tested the performance of 100 random episodes without updating the network. The column  ‘Score’  in the table shows the average score of this setting. The ‘Transition’ indicates the number of transitions (timesteps) that each network has to go through to achieve the same score without BOOK. The 'Ratio' means the ratio of the book size over transition to confirm the sample efficiency of our method. For example, if Dueling DQN learns the BOOK of size 1,000 from A3C in Q*bert, it can get the score of 388.1. If this network learns without BOOK, it has to go through 1,080K transitions. The ratio is 0.09\%, which is 1,000 / 1,080K.

As shown in Table \ref{table:book_exp}, even if RL agents only learn the small-sized BOOK, they can obtain scores similar to those of scores obtained when learning dozen to thousands of times more transitions. In the Cartpole environment, particularly, all models obtained the highest score of 500, except when DQN learn the BOOK written by DQN.

However, the obtained scores are quite different depending on the model that wrote the BOOK and the model that learned the BOOK. In most environments and training models, learning the BOOK written by A3C is better than learning the BOOK written by DQN. Also, even if the same BOOK is used, the performances are different according to the training algorithm. DQN has lower performance than A3C or Dueling DQN in most environments. The major difference in each method is that DQN uses only Q value, and A3C and Dueling DQN use state-value and advantage. Dueling DQN got good scores in most environments, but in the case of Acrobot, using the BOOK by DQN, it was lower than all other models. This indicates that the information stored in the BOOK can be more or less useful depending on the reader RL method.


\begin{figure*}[t]
    \centering
	\subfigure{\includegraphics[width=0.245\linewidth]{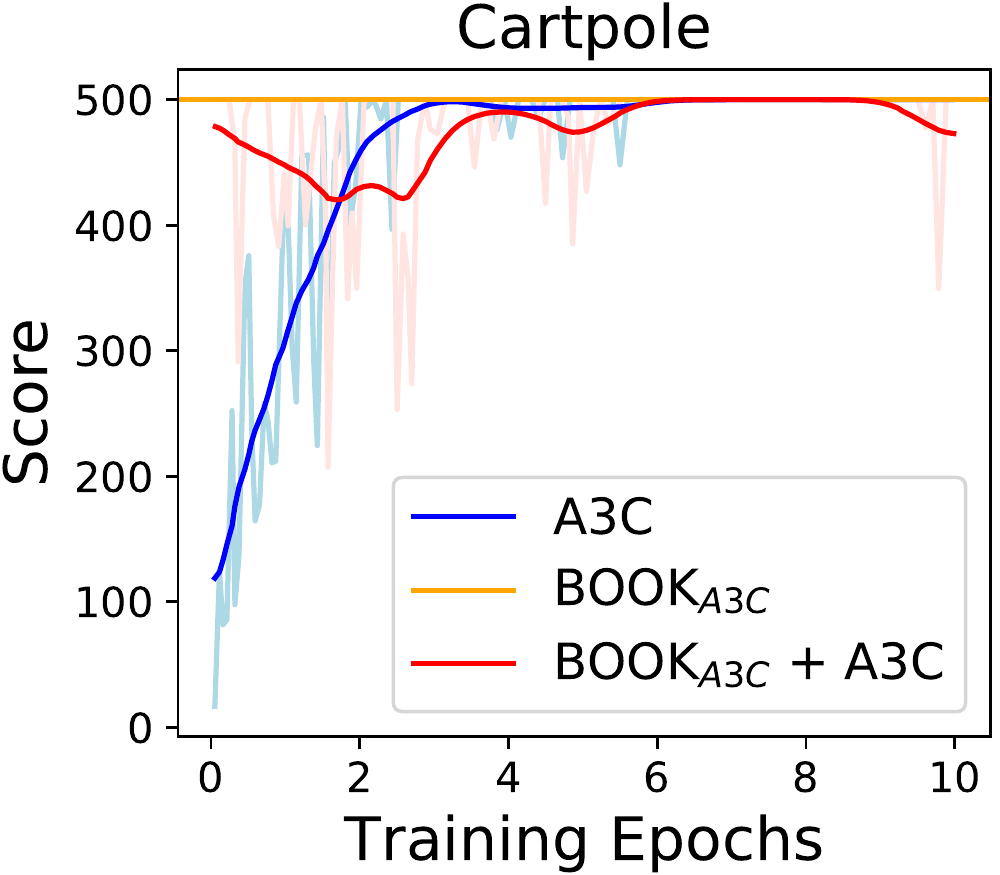}}
    \subfigure{\includegraphics[width=0.245\linewidth]{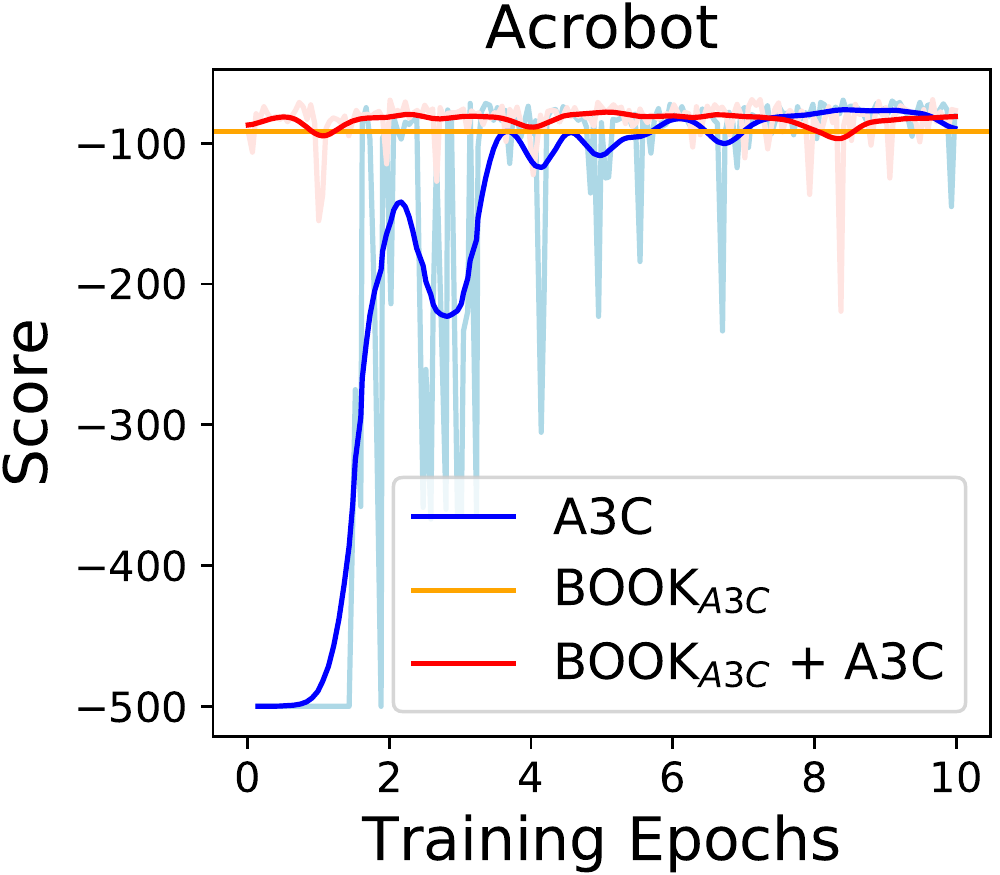}}
    \subfigure{\includegraphics[width=0.245\linewidth]{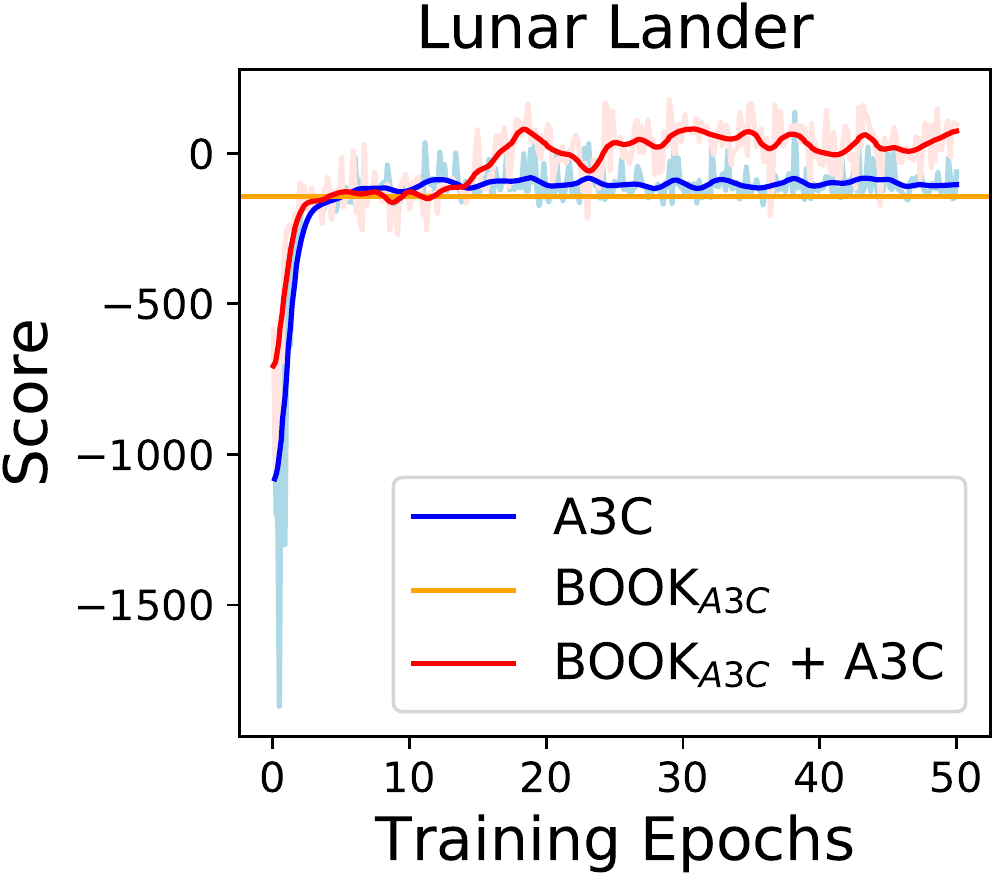}}
    \subfigure{\includegraphics[width=0.245\linewidth]{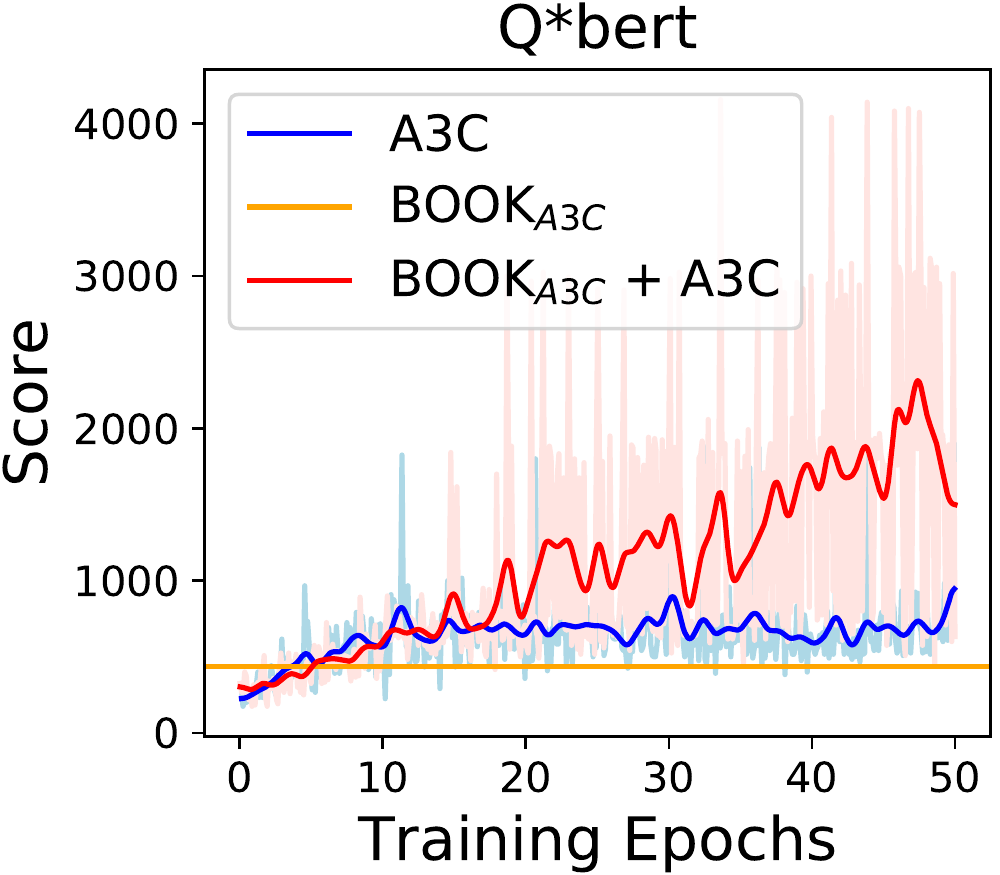}}
	\subfigure{\includegraphics[width=0.245\linewidth]{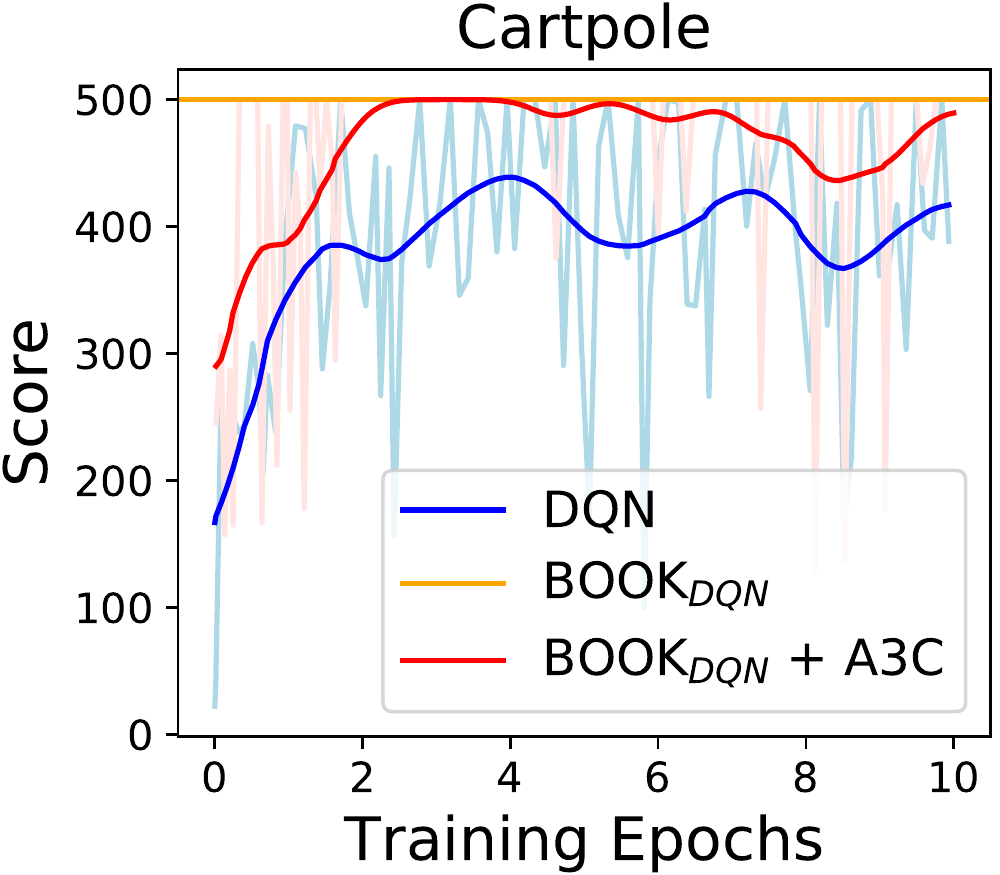}}
    \subfigure{\includegraphics[width=0.245\linewidth]{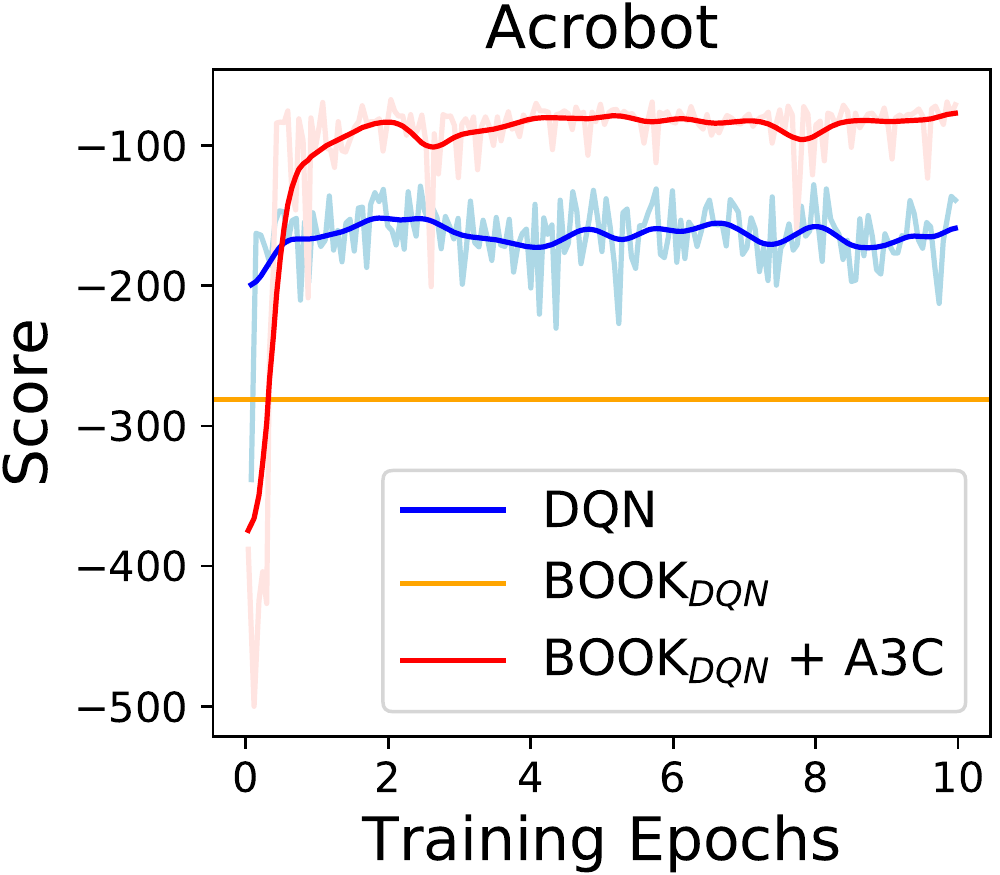}}
    \subfigure{\includegraphics[width=0.245\linewidth]{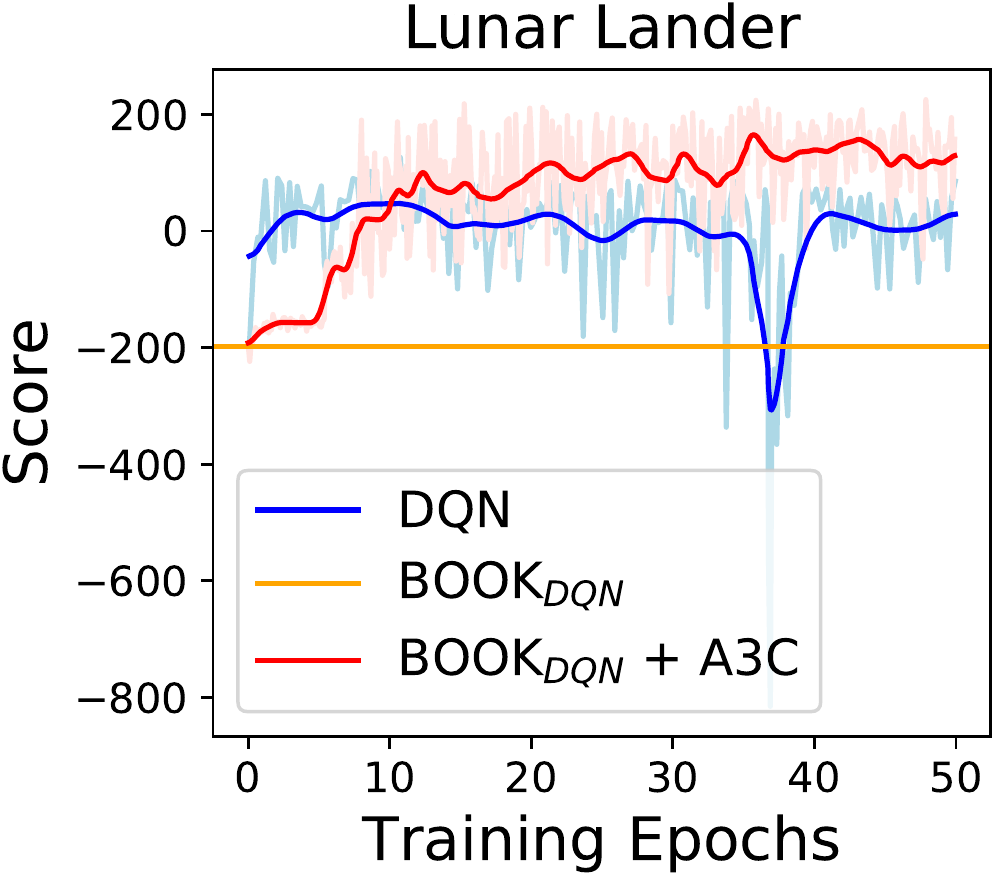}}    
    \subfigure{\includegraphics[width=0.245\linewidth]{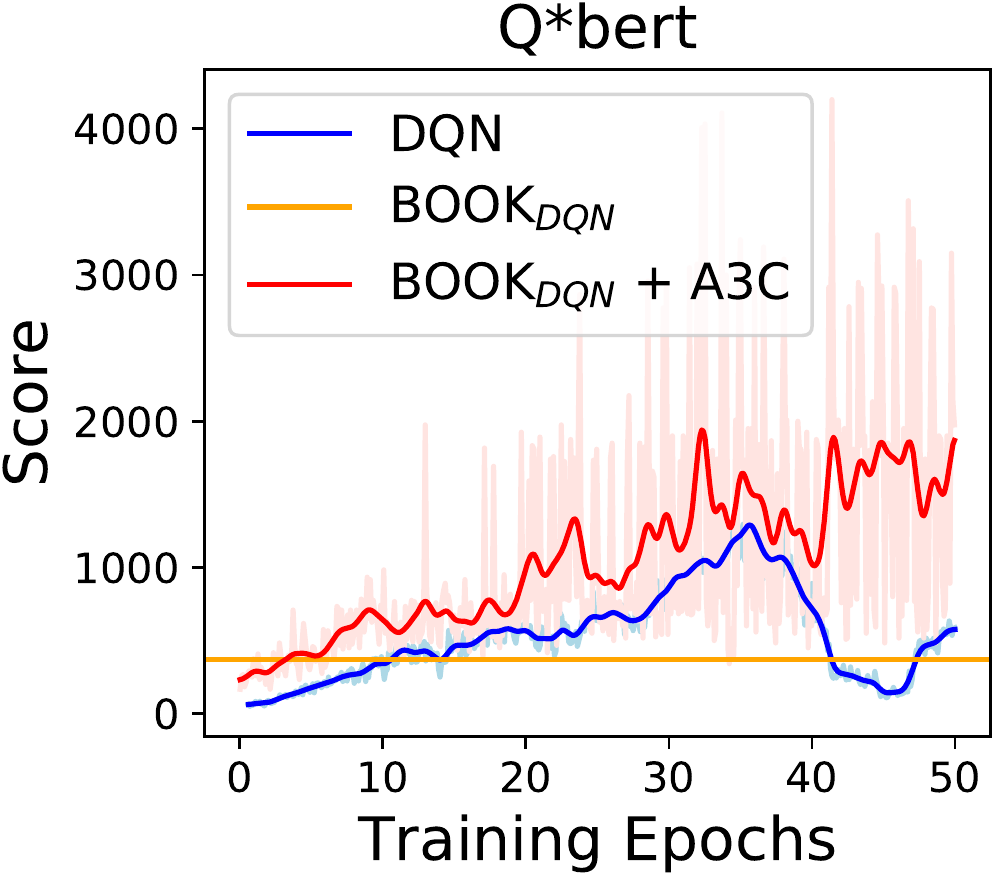}}
    \vspace{-2mm}
    \caption{
Performance for additional training after learning the BOOK. The upper row shows the case where a book is created in A3C and it is trained in a new A3C agent, and the lower row shows the case where a book is created in DQN and trained in A3C.
     \textbf{Blue}: Conventional method (A3C or DQN); 
\textbf{Yellow}: The score of the network that was trained only with the BOOK;
\textbf{Red}: The network was trained using conventional method after learning the BOOK;
    An epoch corresponds to one hundred thousand transitions (across all threads). Light colors represent the raw scores and dark colors are smoothed scores.}
    \label{fig:exp_main}    
\end{figure*}

\textbf{Performance of additional training after learning the BOOK: }
The graphs in Figure~\ref{fig:exp_main} show the performance when the BOOK is used for pre-training the conventional RL networks. After learning the BOOK, each network is trained by each network-specific method.
For this study, we conducted the experiments with two different settings: (1) training the RL network using the BOOK generated by the same learning method, (2) training the RL network using the BOOK generated by the different learning method.
For the first setting, we trained the network and BOOK using A3C~\cite{mnih2016asynchronous},
while in the second, we generated the BOOK using DQN~\cite{mnih2013playing} and trained the network with A3C~\cite{mnih2016asynchronous}.

The results of these two different settings are the upper and the lower rows of Figure~\ref{fig:exp_main}, respectively.
In the upper row, the `blue' line shows the score achieved through training an A3C network from scratch, the `yellow' horizontal line shows the base score which can be achieved through training other A3C network only with a BOOK which is published by a trained A3C network. The `red' line shows the additional training results after training the A3C network with BOOK. In the lower row, the three lines mean the same with the upper row except that the BOOK is published by a different RL network, DQN.

As shown in Figure~\ref{fig:exp_main}, the scores achieved from pre-trained networks using a BOOK were almost the same as the highest scores achieved from conventional methods. Furthermore, additional training on the pre-trained networks was quite effective since they achieved higher scores than conventional methods as training progresses. Especially, BOOK was very powerful when it is applied to a simple environment like Cartpole, which achieved much higher score than conventional training methods. Some experiments show that the maximum score of 'BOOK + A3C' is same with that of 'A3C' but this is because their environments have a limited maximum score. Also, almost every experiments show that the red score starts from lower than the yellow baseline as additional training progresses. It may seem weired but it is very natural phenomenon for the following reasons: (1) As additional training begins, exploration is performed. (2) BOOK stores Q value with actual reward without bootstrapping, but DQN and A3C use bootstrapped Q value, thus they (actual and bootstrapped Q-values)  don't match exactly.



\subsection{Qualitative Analysis}
To further investigate the characteristics of the proposed method, we conducted some experiments by changing the hyper-parameters.

\begin{table}[t]
\vskip -0.08in
\caption{Average scores that can be obtained by learning a BOOK of a certain size and the number of transitions that is needed for A3C to get the same score. 
}
\label{table:book_size} 
\begin{center}
\begin{small}
\begin{sc}
\resizebox{0.99\linewidth}{!}{
\begin{tabular}{@{\hskip2pt}c|c@{\hskip5pt}c|c@{\hskip5pt}c|c@{\hskip5pt}c@{\hskip2pt}}
\toprule
 & \multicolumn{2}{c|}{Cartpole} & \multicolumn{2}{c|}{Acrobot} & \multicolumn{2}{c}{Q*bert}\\  
\midrule
Size&Score&Transition&Score&Transition&Score&Transition\\
\midrule
250 & 114.0 & 25.8K& -143.8 &330K &231.6 & 78K\\
500 &\textbf{500.0} & \textbf{324K} &-158.5 & 204K &371.8&271K\\
1000 &500.0 &324K &\textbf{-91.8} & \textbf{372K}& 436.0 &383K\\
2000 &500.0 &324K &-93.2 &363K & \textbf{520.0}&\textbf{618K}\\
\bottomrule
\end{tabular} 
}
\end{sc}
\end{small}
\end{center}
\vskip -0.1in
\end{table}

\begin{figure}
\centering
\subfigure[Quantization Level]{\includegraphics[height=2.7cm]{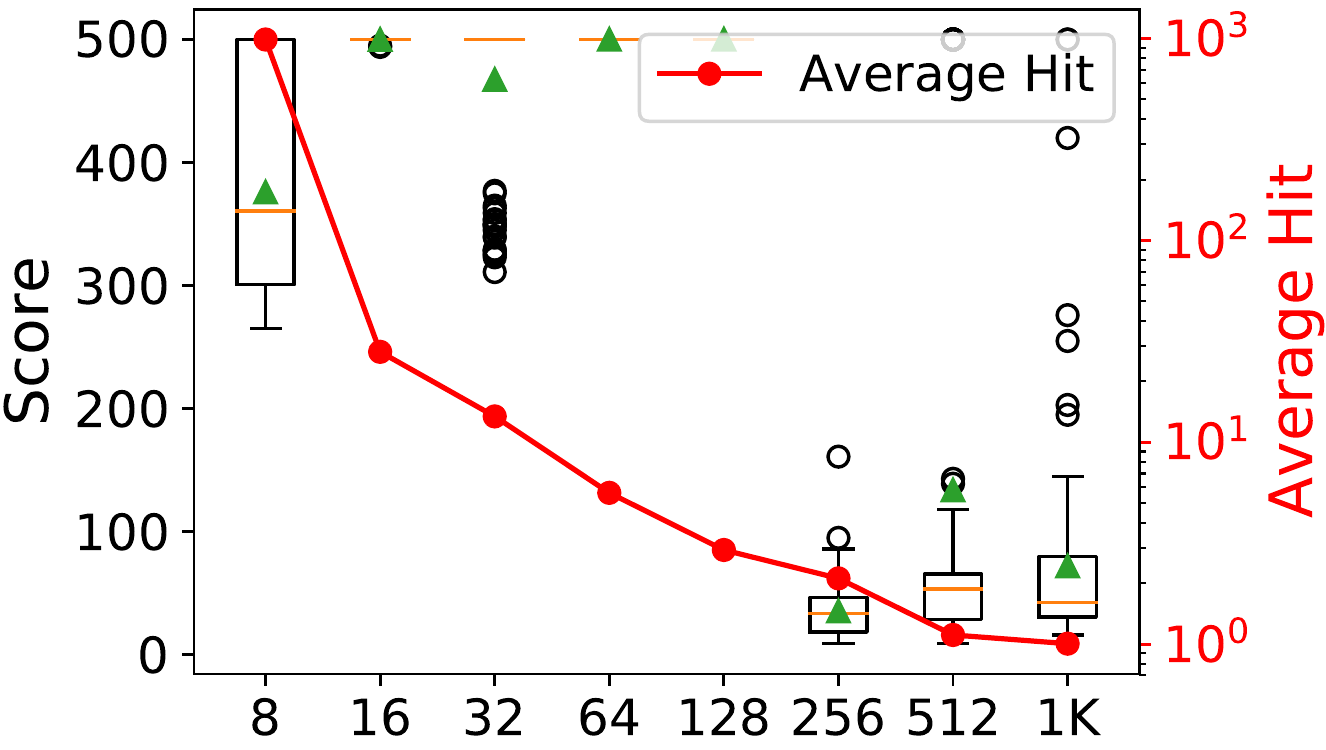}}
\subfigure[Priority Methods]{\includegraphics[height=2.7cm]{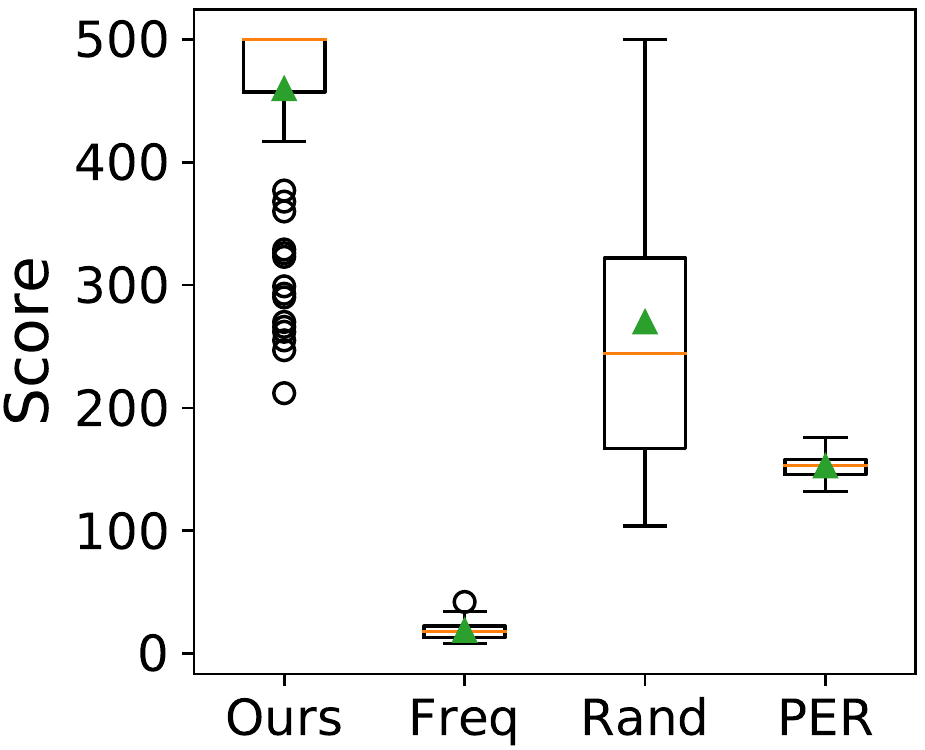}}
\vspace{-3mm}
\caption{(a) Distribution of scores according to the quantization level and the average number of hits in each Linguistic State $c_k$ included in the BOOK. (b) Distribution of scores when we use three different methods than our proposed priority method. \textbf{Freq}: state visiting frequency, \textbf{Rand}: random state selection, \textbf{PER}: priority term from \textit{prioritized experience replay}. All data were tested in Cartpole, and scores were measured in 100 random episodes. The green triangle and the red bar indicate the mean and the median scores, respectively. Blank circles are outliers. }
\label{fig:exp_method}
\vskip -0.1in
\end{figure}

\noindent
\textbf{Learning with different sizes of BOOKs: }
%
To investigate the effect of the BOOK size, we tested the performance of the proposed method using the published BOOK size of 250, 500, 1000, and 2000.
Table \ref{table:book_size} shows the score obtained by our baseline network which was trained using only the BOOK in a specified size.
Also, in the table, we showed the number of transitions (experiences) that a conventional A3C has to go through to achieve the same score. 
This result shows that a relatively small number of linguistic states can achieve a score similar to that of the conventional network with only the published BOOK. As shown in the table, training an agent in a complex environment requires more information and therefore a larger BOOK is needed.
\noindent
\textbf{Effects of different quantization levels:}
%
In this experiment, we confirmed the performance difference according to the resolutions of linguistic function.
First of all, we differentiated the quantization level and published a BOOK of 1,000 size to check the difference of performance according to the resolutions of linguistic function.  Figure \ref{fig:exp_method}(a) shows the distribution of scores according to the quantization level (quartile bar) and the average number of hits in each linguistic state $c_k$ included in the BOOK (red line).


From Fig. \ref{fig:exp_method}(a), we found that the number of hit for each linguistic state decreases exponentially as the quantization level increases. 
Also, when the quantization level is high, the importance of $c_k$ in equation (\ref{eq:importance}) couldn't be defined and its score decreased because hit ratio becomes low. It can be seen that the highest and stable scores are obtained at quantization level of 64 and 128.

\textbf{Comparison of the priority methods: }
Also, to verify the usefulness of our priority method (\ref{eq:priority}), 
we tested the algorithm with different design of the priority; random selection, frequency only, method from \textit{prioritized experience replay}~\cite{schaul2015prioritized}, and the proposed priority method. A book capacity $K$ was set to 10,000 for this test.

As shown in  Figure~\ref{fig:exp_method}(b), the algorithm applying the proposed priority term achieved clearly far superior performance than other settings.
We note that the case of using only frequency term marked the lowest performance, even lower than the random case.
This is because the learning process proceeds only with the experiences that appear frequently when the priority is set only by the frequency. 
Correspondingly, the information of the critical, but rarely occurred experiences are not reflected enough to the training and hence, leads to inferior performance.

The priority term of the \textit{prioritized experience replay} also marked poor results. It is better than using frequency only as a priority, but even lower than random selection. This algorithm is intended to give priority to the states that are not yet well learned among the entire experience replay memory, \sm{and is not designed to extract a few core states.}

\subsection{Implementation Detail}
We set the maximum capacity $K$ 
of a BOOK to $100,000$ while writing the BOOK.
To maintain the size of the BOOK, only the top $50$\% experiences are preserved and the remaining experiences are deleted to save new experiences when the capacity exceeds $K$.
As a linguistic rule, each dimension of the input state was quantized into $128$ levels. We set the discount factor $\gamma$ for rewards to $0.99$. Immediate reward $r$ was clipped from $-1$ to $1$ at Q*bert and generalized with $\tanh(r/10)$ for the other 3 environments (Cartpole, Acrobot and Lunar Lander). The frequency limit $F_l$ was set to $20$ and the decay period $T$ was set to $100$.

Our method adopted the same network architecture with A3C for Atari Q*bert. But for the other 3 environments, we replaced the convolution layers of A3C to one fully connected layer with 64 units followed by ReLU activation.
Each environment was randomly initialized. For Q*bert, it skipped a maximum of 30 initial frames for random initialization as in \cite{dmitry2016async}.
We used 8 threads to train A3C network and instead of using shared RMSProp, ADAM~\cite{kingma2014adam} optimizer was used. All the learning rates used in our experiments were set to $5\times 10^{-4}$.
To write a BOOK, we trained only $1$ million steps (experiences) for Cartpole and Acrobot and $5$ million steps for Lunar Lander and Q*bert.
After publishing a BOOK, we pre-trained a randomly initialized network for $10,000$ iterations with batch size $8$, using only the contents in the published BOOK. It took less than a minute to learn a BOOK with 1 thread on Nvidia Titan X (Pascal) GPU and 4 CPU cores, for Q*bert.

\section{Conclusion }

In this paper, we have proposed a memory structure called BOOK that enables sharing knowledge among different deep RL agents. 
Experiments on multiple environments show that our method can achieve a high score by learning a small number of core experiences collected by each RL method. It is also shown that the knowledge contained in the BOOK can be effectively shared between different RL algorithms, which implies that the new RL agent does not have to repeat the same trial and error in the learning process and that the knowledge gained during learning can be kept in the form of a record.

\sm{\nj{As future works}, we intend to apply our method to the environments with \nj{a} continuous action space. \nj{Linguistic functions can also} be defined in other ways, such as neural networks, for better clustering and feature representation.}



\nocite{langley00}

\bibliography{example_paper}
\bibliographystyle{icml2018}

\end{document}